\documentclass[12pt,a4paper]{article}
\usepackage{graphicx}
\usepackage{amsfonts} 
\usepackage{amssymb}
\usepackage{amsmath}
\usepackage[a4paper,margin=4cm,headheight = 15pt]{geometry}
\usepackage{lastpage}
\usepackage{fancyhdr}
\usepackage{natbib}	
\usepackage{hyperref}
\usepackage{multicol}
\usepackage{url}
\usepackage{listings}
\usepackage[T1]{fontenc}
\usepackage{float}
\usepackage{caption}
\usepackage{verbatim}
\usepackage{subcaption}
\renewcommand\harvardurl[1]{\textbf{URL}: \url{#1}} 
\usepackage{xurl}

\newcommand{\titlestr}{Investigating the Impact of 9/11 on \textit{The Simpsons} through Natural Language Processing}
\newcommand{\shorttitlestr}{\textit{The Simpsons} and 9/11}
\newcommand{\authorstr}{Athena Xiourouppa} 

\begin{document}
\begin{titlepage}
  \centering
  
  {\LARGE \titlestr \par}

  \vspace{1cm}
  {\Large \authorstr \par}

  {\bf a1764132}

  \vspace{1cm}
  October 19th, 2021     

\includegraphics[width=0.35\textwidth]{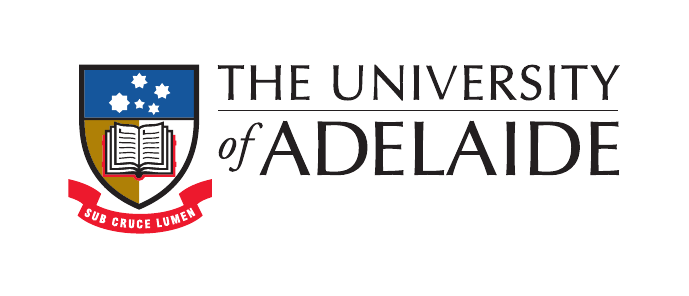}

  \vspace{2cm}
  \flushleft
  Project Area: {\bf Natural Language Processing} \\
  Project Supervisor: {\bf Matt Roughan} \\

  \vspace{5mm} {\footnotesize In submitting this work I am indicating
    that I have read the University's Academic Honesty Policy. I
    declare that all material in this assessment is my own work except
    where there is clear acknowledgement and reference to the work of
    others.\par}

  \vspace{5mm} {\footnotesize I give permission for this work
    to be reproduced and submitted to other academic staff for
    educational purposes.\par}

  \vspace{5mm} {\footnotesize {\bf OPTIONAL:} I give permission this work
    to be reproduced and provided to future students as an exemplar report.\par}

  \vfill
\end{titlepage}

\pagestyle{fancy}
\fancyhf{}
\rhead{\shorttitlestr}
\lhead{\authorstr}
\rfoot{Page \thepage\ of \pageref{LastPage}}
\renewcommand{\headrulewidth}{1pt}

\begin{abstract}
The impact of real-world events on fictional media is particularly apparent in the American cartoon series \textit{The Simpsons}. While there are often very direct pop-culture references evident in the dialogue and visual gags of the show, subtle changes in tone or sentiment may not be so obvious. Our aim was to use Natural Language Processing to attempt to search for changes in word frequency, topic, and sentiment before and after the September 11 terrorist attacks in New York. No clear trend change was seen, there was a slight decrease in the average sentiment over time around the relevant $2000-2002$ period, but the scripts still maintained an overall positive value, indicating that the comedic nature of \textit{The Simpsons} did not wane particularly significantly. The exploration of other social issues and even character-specific statistics is needed to bolster the findings here. 
\end{abstract}

\clearpage
\section{Introduction}
Popular television series, \textit{The Simpsons}, is currently America's longest-running animated series, and America's longest-running scripted prime-time television series, according to \cite{keveney_2019}. \textit{The Simpsons} also possesses an array of other accolades earning it a spot in \cite{guiness_2021} Hall of Fame. Known for its satirical depiction of the middle-class, nuclear American family and pop-culture references, sometimes even before an event actually occurs, it is no surprise the series has maintained its longevity.

However, what is of interest here is not only the impact of general world events on the production of the show, but in particular, those with a level of emotional or even traumatic impact. What often comes to mind as a significant effect in recent American history are the September 11, 2001 terrorist attacks, commonly referred to as 9/11. Even two decades later, the event is still widely regarded with sombre sensitivity, \cite{hajjar_2021} noting that its alleged perpetrators still await trial. The event sparked a rapid change in American foreign policy, including the War on Terror.

While \textit{The Simpsons}, and other television shows, be they sitcom or drama, often feel free to poke fun or make light of major events, 9/11 was not given the same treatment. In fact, it was treated with a level of severity and delicacy not generally seen before. \cite{enwiki:1050053021} even compiles a list of major changes made to various forms of media, from digitally removing images of the World Trade Center, to even reanimating an entire sequence of a childrens' animated film, as shown in \cite{vox_2017}. However, there may still be deeper, more nuanced changes in media that are not immediately made obvious.

In order to analyse the potential impact of 9/11 on \textit{The Simpsons} with a level of not only objectivity, but also subtlety, a simple viewing is not sufficient. Natural Language Processing is able to look at mathematical patterns and trends within the entire script corpus. Some particular analyses that are helpful here include frequency distribution and sentiment analysis. The aim is to find any possible fluctuation in the frequency of words with a relation to 9/11. Determining the right words to consider is its own task, and requires the use of term frequency–inverse document frequency, a frequency algorithm that compares two corpora in order to determine a word's relative frequency or relevance to a topic. The sentiment change looks at the emotional connotations of dialogue, by scoring words as positive, neutral, or negative. If there is a notable feature of the episodes which were aired around the $2001-2002$ period, accounting for the time delay between an episode's writing and airing, this would likely be connected to 9/11. These analyses are also useful to determine statistical differences between spoken dialogue and more standard text, particularly whether the strong use of colloquial language is  impactful on the distributions obtained.

Results generally showed no conclusive evidence of a distinct change before and after $9/11$ in the dialogue, however many improvements are possible. There was a slight decrease in average sentiment over time, but there was no clear trend in changes of word frequencies, regardless of the method used. A large total of $75$ topics were deemed optimal in unsupervised classification. However, there are many limitations in the process which could be causing this such as the stemmer removing proper nouns, and the lack of consideration of bigrams and trigrams, which provide context to words.

\clearpage
\section{Background and Methodologies}
All relevant Python notebooks are linked in the following Github repository, \href{https://github.com/athenax-coder/NLP-The-Simpsons-and-9-11}{NLP-The-Simpsons-and-9-11}.
\subsection{Data Cleaning}
Six hundred episodes of script data from \textit{The Simpsons}, including dialogue and background details such as scene location, over the period $1989-2016$ were obtained from \cite{banerjee_2019}. The relevant file here is \texttt{simpsons\_script\_lines.csv}, a comma-separated value file containing not only the script text, but other details such as the corresponding character, timestamp, episode ID, and word count. There are $158271$ rows, \textit{i.e.,} lines within the script.

The raw text was manually filtered to maintain spoken lines, maintaining both dialogue and unspoken expressions. For example, the catchphrase of the protagonist, Homer Simpson, `D'oh' is written as (ANNOYED GRUNT). All other columns beside the raw text and index were subsequently removed.

All words were converted to lowercase, so that the frequency counter would not consider the same word as separate cases. Each line of raw text also had the speaker designation within it, denoted by the character's name, a colon, then followed by the line, so everything prefacing the colon was removed. Other grammar, and the leading white-space that resulted was also removed.

This cleaned data file was exported as both a comma-separated value file and a text file. This is because some of the later analyses work within a data frame, while others need the corpus as a single string. 

In order to partition the data frame by year, a similar cleaning process was used, however the ID of the episode was also kept. Another file provided by \cite{banerjee_2019} was \texttt{simpsons\_episodes.csv} which contains data about the episodes themselves, including title, air date, and IMDB rating. The air year was extracted and then appended to the original cleaned data frame, ensuring the indices matched. This allowed the creation of a dictionary to map episode IDs to air year, and therefore allow the dialogue for a particular year or years to be extracted from the data frame.

Some analyses required the input string to be tokenized, that is, parsed so each individual word can be identified. This essentially partitions the string by white-space and punctuation. Once this was done, a decision was made to remove stop words, using the default list from the Python library \textit{NLTK}. Stop words are commonly used words that do not add meaning to a text, so in order to look for changes in sentiment, or even compare the frequency of special words, such as those related to 9/11, the noise created by stop words needed to be removed.

\subsection{Exploratory Analysis}
To find the most commonly spoken words, we created a frequency distribution for the cleaned and tokenized data. Initially, the stop words were kept, but this did not provide meaningful results, as these stop words were the most commonly used anyway. We also calculated the lexical diversity, that is, the variety of different words in the text.

The frequencies were used to calculate Zipf's Law, a statistical relationship between the frequency of word usage and the number of words in a text. \cite{debowski_2000} explains that on average, the frequency of a word token $w$, $f$, is inversely proportional to its rank, $r$, shown in \eqref{zipf}.
\begin{equation}
    f(w) \: \propto \: \frac{1}{r(w)^{\alpha}}.
    \label{zipf}
\end{equation}
The estimation of a word's probability of occurring is shown in \eqref{zipfprob}, where $N$ is the total number of tokens.
\begin{equation}
    p(w) = \frac{f(w)}{N}.
    \label{zipfprob}
\end{equation}
Another statistical measure of a text's diversity is its entropy, or measure of randomness. The equation for entropy is shown in \eqref{entropy}, which is a function of the probability of each word, $w$, appearing in the corpus, $W$. \cite{dan_poddar_ganesan_2012} states that a higher entropy is indicative of less predictability in text, which is also associated with building language models using a corpus to generate text.
\begin{equation}
    H(w) = -\sum_{w \in W} p(w)\log_2p(x).
    \label{entropy}
\end{equation}
Finally, Heaps' Law refers to the proposed power relationship between the number of unique words, $V$ in a corpus, compared to the total number of words, $n$ in the corpus.
\begin{equation}
    V(n) = Kn^{\beta}.
    \label{heaps}
\end{equation}
In \eqref{heaps}, $K$ and $\beta$ are empirically-determined parameters, with \cite{cook_2019} using values of $K$ between $10$ and $100$ and \cite{chacoma_zanette_2020} using $\beta \approx 0.6$. Generally, English texts tend to exhibit asymptotic behaviour for large $n$.

\subsection{Sentiment Analysis}
\cite{farhadloo} describe sentiment analysis as the process of identifying the underlying opinion or emotion within text. This is no trivial task and often requires extensive machine learning in order to effectively train an algorithm to correctly classify words. Words are typically classified as being positive, neutral, or negative, however this understandably depends on the context, \textit{i.e.,} the surrounding words in a sentence. Factors such as sarcasm also contribute to the difficulty of the task. Two analysers in Python were compared, from both the \textit{TextBlob} and \textit{VADER} packages. 

\subsection{Topic Analysis}
\subsubsection{Latent Dirichlet Allocation}
Grouping text into unsupervised topics can give insight into the distributions and common themes within a text. \cite{blei_ng_jordan_2003} describes the process of Latent Dirichlet Allocation, which is a probabilistic model used to analyse the corpus of a text and partition words into a predefined number of topics. 

Defining the number of topics is often done based on a coherence score, which \cite{kapalia_2019} states is based on the extent of semantic, that is, dictionary meaning similarity between the most frequent words in a single topic.

\subsubsection{Supervised Allocation Using TF-IDF}
When the topic of interest is known, one may use the Term Frequency Inverse Document Frequency or TF-IDF to find words most commonly associated with a topic. In particular, the TF-IDF is more rigorous than a standard frequency calculation, because it assesses whether the word is particularly prominent or special to that document. \cite{ramos_2003} mentions the use of a subset of a large corpus, however this process can also be done by comparing the corpus of interest with a more generic corpus in the same language.

Once a short list of words strongly associated with a topic, in this case, 9/11, is established, one can return to the original corpus of interest and search for the corresponding frequencies. This can give an indication of how often those words appear in the text, as opposed to a standard frequency method which may create noise when words are used synonymously or commonly.

\clearpage
\section{Results}
\subsection{Exploratory Analysis}
\begin{figure}[H]
    \centering
    \includegraphics[width=\textwidth]{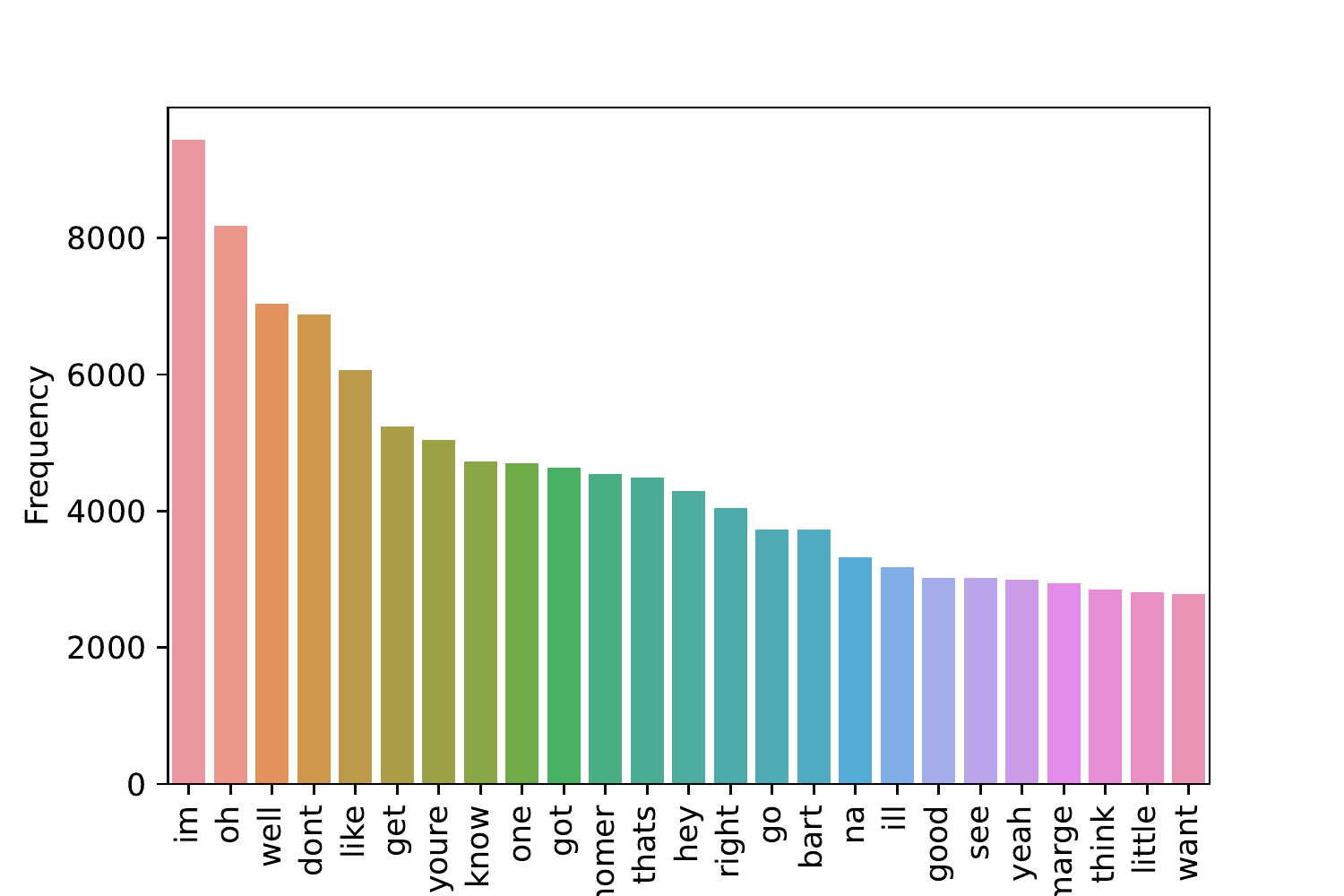}
    \caption{The top twenty-five words spoken in \textit{The Simpsons} plotted by decreasing frequency.}
    \label{top25}
\end{figure}

\begin{figure}[H]
    \centering
    \includegraphics[width=\textwidth]{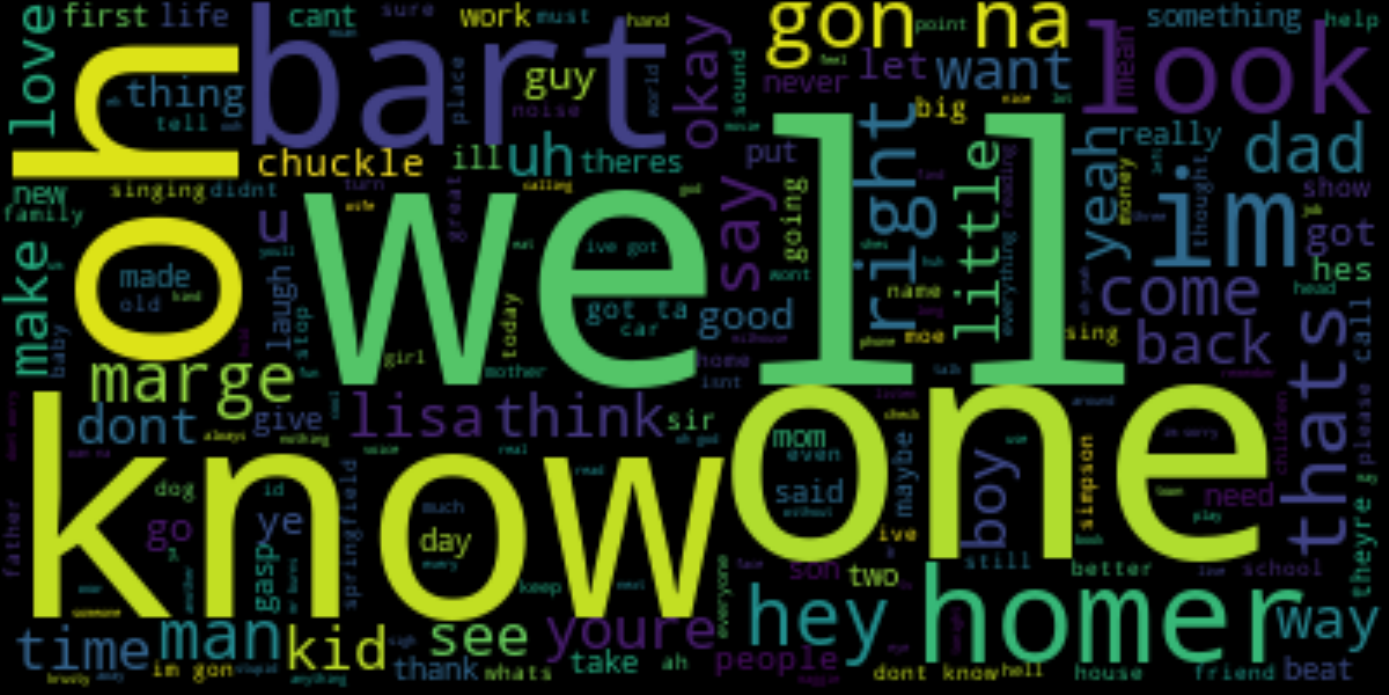}
    \caption{Word cloud of \textit{The Simpsons} with more frequently spoken words appearing larger.}
    \label{wc_simp}
\end{figure}

Figure \ref{top25} shows a graph of the top twenty-five spoken words in \textit{The Simpsons} by frequency. We note the common occurrence of the names of main characters Homer and Marge. Some other words in the list such as `im', `oh', and `hey' are somewhat colloquial, and typical of dialogue. Figure \ref{wc_simp} visualises this in a word cloud, which allows other character names such as Bart to be seen. The lexical diversity was around $6\%$, which is expected to be lower than when the stop words are contained.

Another analysis performed with the partitioned data was identifying the most common word per year, with results shown in Table \ref{wordperyear}. The word `thats' appears in $1994$, $1996$, and $2004$, although being a common speech phrase, this is not particularly notable. There are however some abnormalities, such as `camel' and `pony' in $2000$ and $2001$ respectively.

\begin{table}[H]
\centering
\begin{tabular}{|c|c|}
\hline
\textbf{Year} & \textbf{Word} \\ \hline
1989          & `homer'       \\ \hline
1990          & `come'        \\ \hline
1991          & `calling'     \\ \hline
1992          & `money'      \\ \hline
1993          & `homer'       \\ \hline
1994          & `thats'       \\ \hline
1995          & `like'        \\ \hline
1996          & `thats'       \\ \hline
1997          & `laughs'      \\ \hline
1998          & `waiting'     \\ \hline
1999          & `whatchu'     \\ \hline
2000          & `camel'       \\ \hline
2001          & `pony'        \\ \hline
2002          & `accustomed'  \\ \hline
2003          & `sing'        \\ \hline
2004          & `thats'       \\ \hline
2005          & `party'       \\ \hline
2006          & `grumply'     \\ \hline
2007          & `arent'       \\ \hline
2008          & `urgent'      \\ \hline
2009          & `youre'       \\ \hline
2010          & `love'        \\ \hline
2011          & `dodgeball'   \\ \hline
2012          & `grumpy'      \\ \hline
2013          & `learn'       \\ \hline
2014          & `birth'       \\ \hline
2015          & `monotone'    \\ \hline
2016          & `series'      \\ \hline
\end{tabular}
\caption{The most frequent word per year from $1989$-$2016$ in \textit{The Simpsons}.}
\label{wordperyear}
\end{table}

\begin{figure}[H]
    \centering
    \includegraphics[width=\textwidth]{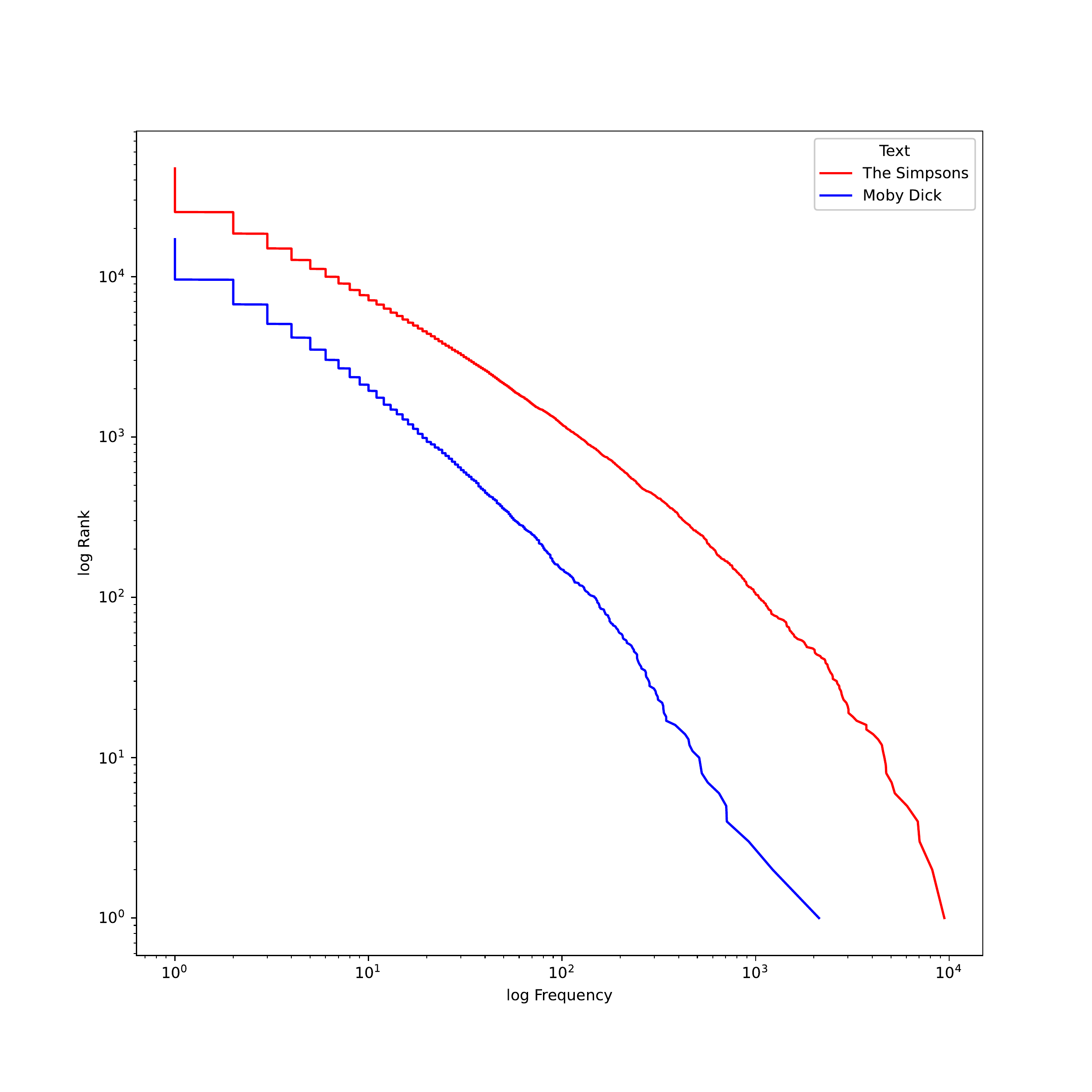}
    \caption{Log-Log plot of word frequency for \textit{The Simpsons} and \textit{Moby Dick}, noting that although there is non-linearity in both texts, \textit{The Simpsons} appears to have a higher degree of randomness.}
    \label{zipfplot}
\end{figure}

To investigate whether \textit{The Simpsons} conforms to Zipf's Law, the relevant calculations were performed and compared against text from the novel \textit{Moby Dick}, which had also been cleaned in a similar fashion. Figure \ref{zipfplot} shows the comparison of the logarithmic plot between frequency and rank. The expectation is a linear relationship between frequency and rank, but notably both have curvature as frequency increases.

Table \ref{entropytab} shows the comparison of entropy values for each text. Both are close in value, indicating a similar degree of randomness.
\begin{table}[H]
    \centering
    \begin{tabular}{|c|c|}
    \hline
    \textbf{Text}         & \textbf{Entropy} \\ \hline
    \textit{The Simpsons} & 11.793           \\ \hline
    \textit{Moby Dick}    & 11.921                 \\ \hline
    \end{tabular}
    \caption{Comparison of entropy values between texts.}
\label{entropytab}
\end{table}

\begin{figure}[H]
    \centering
    \includegraphics[width=\textwidth]{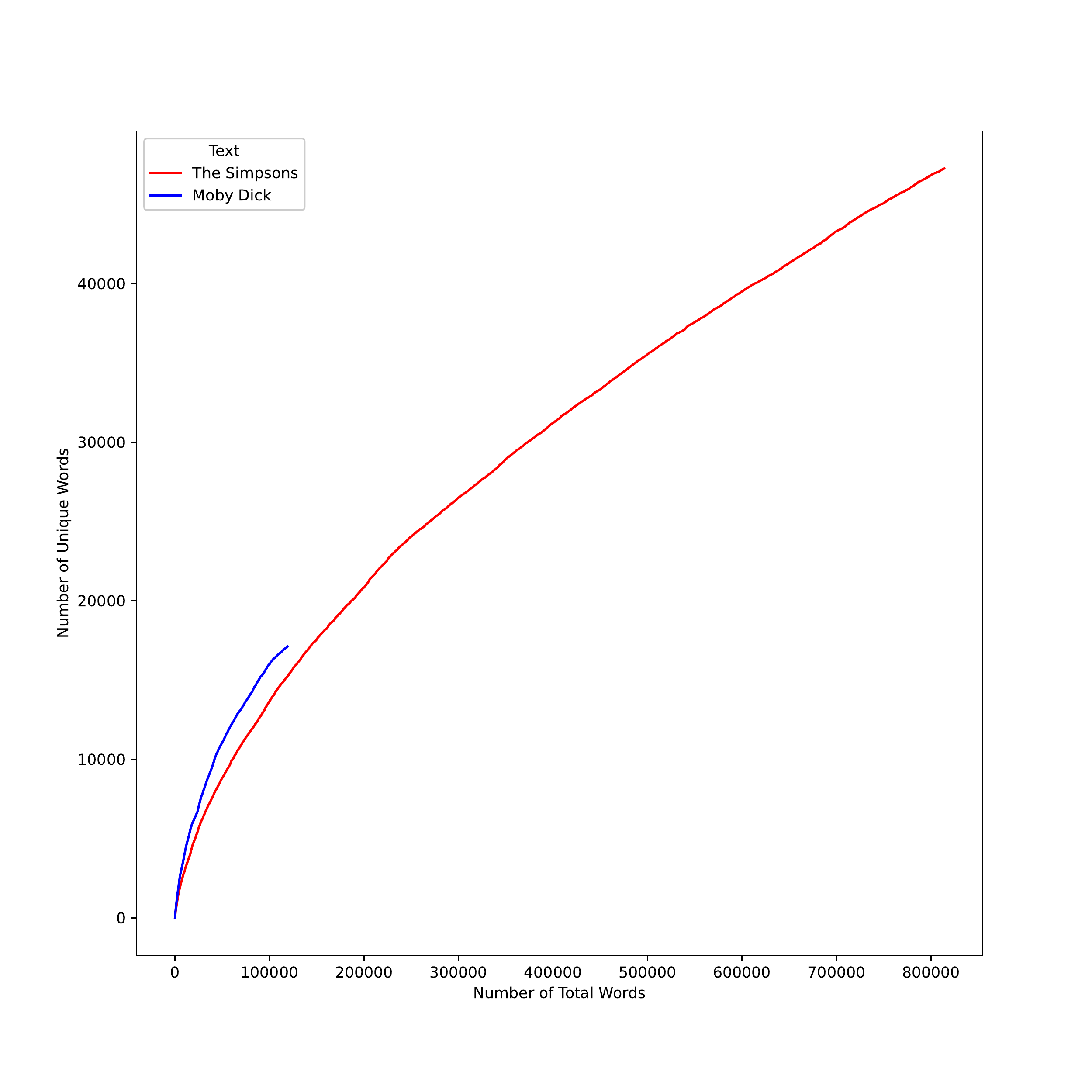}
    \caption{Plot of Heaps' Law for \textit{The Simpsons} and \textit{Moby Dick}, noting that although there is a distinct size difference, both seem to show continuous growth in vocabulary diversity as the total number of words increases.}
    \label{heaplot}
\end{figure}

Finally, Figure \ref{heaplot} shows the plot of predicted unique words against total words for both of the texts. Note the tremendous size difference between the two vocabularies, and that the typical asymptotic behaviour seen for English texts is not particularly evident here. This pertains to many of the specific words invented for \textit{The Simpsons}, including the plethora of character names over the series' run.

\subsection{Sentiment Analysis}
In this case, there are two sentiment analysers within the Python libraries \textit{TextBlob} and \textit{NLTK}. Both were used to compare the average sentiment of each line of \textit{The Simpsons} over time. 

\textit{TextBlob} scored the polarity of each dialogue line over the domain $[-1, 1]$, with $1$ indicating a strongly positive sentence, $-1$ indicating a strongly negative sentence, and $0$ indicating a neutral sentence. The mean sentiment was $0.04613$, indicating a slighly positive bias. Many lines managed to achieve the minimum and maximum values, for example, ``hes planning something evil i know it it must have something to do with the towns water supply" scored $-1$, most likely for the use of the word `evil', whilst ``marge this is the greatest gift any wife has ever given her husband" scored $1$, possibly due to `greatest'.

The VADER (Valence Aware Dictionary and sEntiment Reasoner) sentiment analyzer from \textit{NLTK} showed more subtleties in its calculations, as it not only displayed the proportions of positive, negative, and neutral words per line, but calculated a value to a higher level of resolution of around $4$ decimal places. This is also likely because the analyzer was specifically engineered for the colloquialism of social media, which is understandably suitable for the typically casual, comedic dialogue in \textit{The Simpsons}. The most negative sentence was ``singsongy hell hell hell hell hell hell hell hell hell hell hell hell hell" and scored $-0.9966$, whilst the most positive sentence was ``singing they love  they share  they share and love and share  love love love  share share share  the itchy scratchy showwww" and scored $0.9850$.

\begin{figure}[H]
    \centering
    \includegraphics[width=\textwidth]{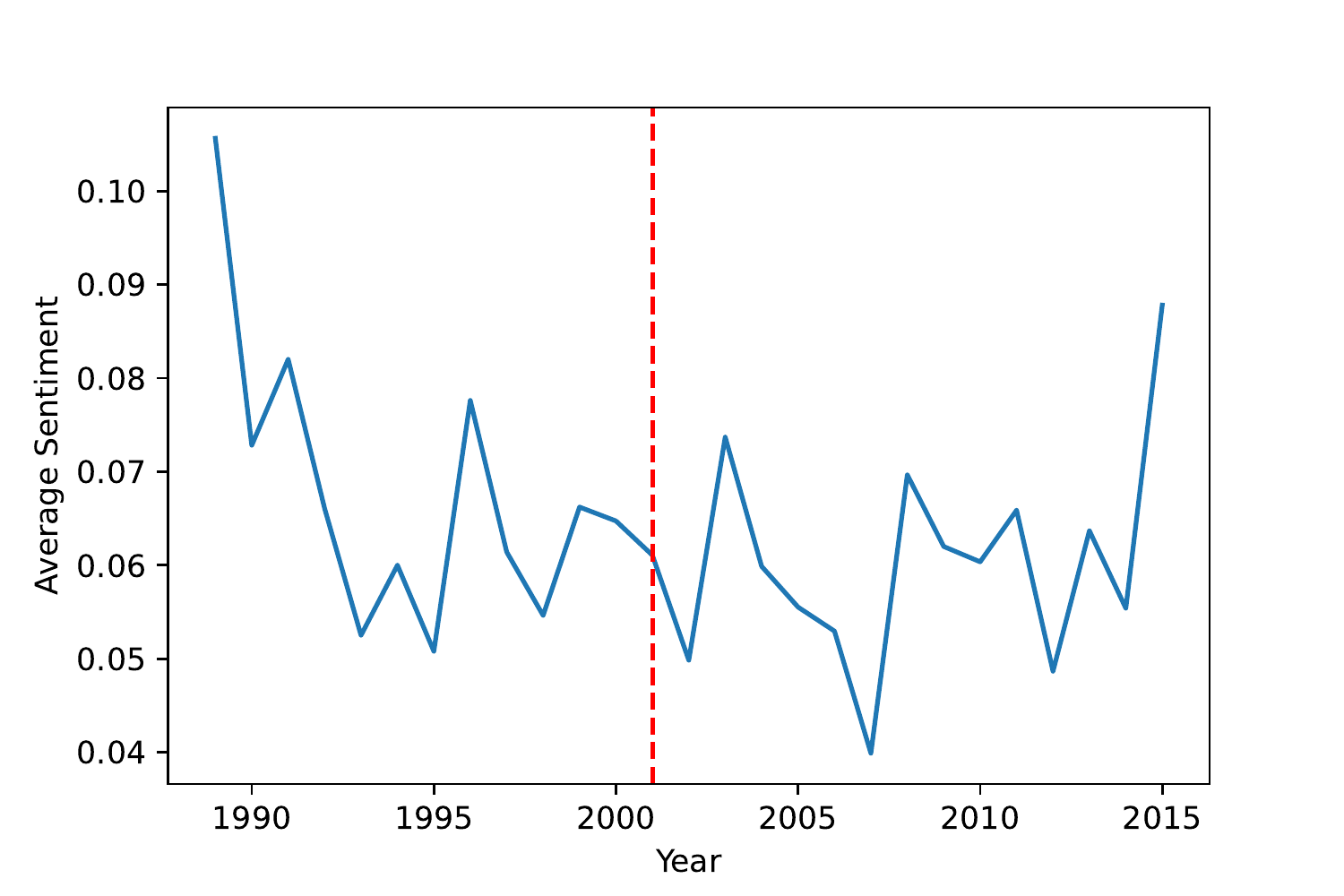}
    \caption{The average sentiment of \textit{The Simpsons} dialogue per year from $1989-2016$. $2001$ is indicated by the red, dashed line.}
    \label{sentyear}
\end{figure}

Figure \ref{sentyear} shows the average sentiment per year, interestingly noting the general decreasing trend, but also that the average sentiment is always strictly greater than zero, indicating at least some emphasis of positive speech. This is understandable considering that \textit{The Simpsons} is primarily designed to be comedic, so real-world events aside, it would be unlikely to see any particularly dramatic drops.

\subsection{Topic Analysis}
\subsubsection{Latent Dirichlet Allocation}
\begin{figure}[H]
    \centering
    \includegraphics[width=\textwidth]{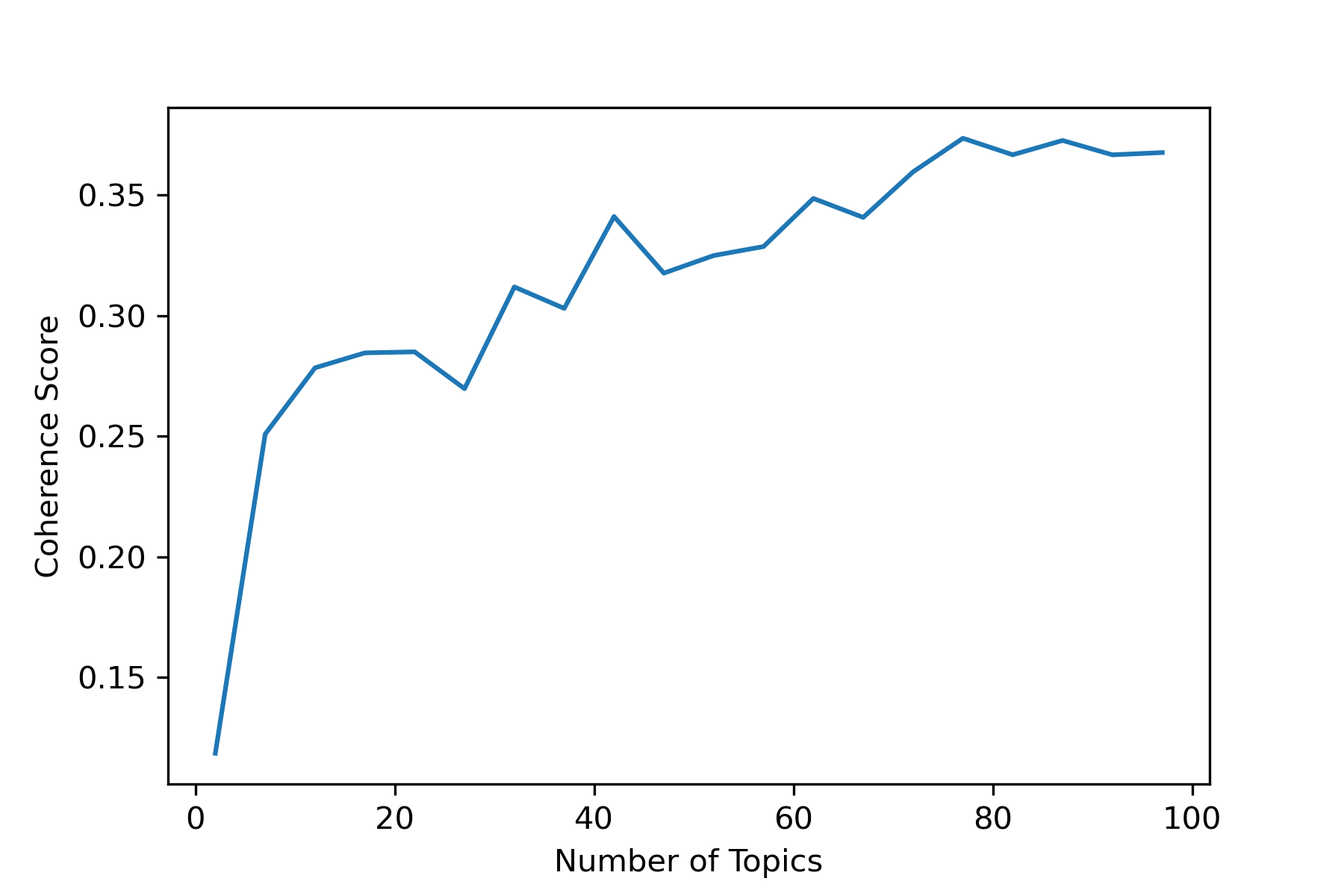}
    \caption{The coherence score of \textit{The Simpsons} for a range of numbers of topics, noting the general plateau at around $75$ for a score of roughly $0.37$.}
    \label{cohscore}
\end{figure}

To compute the optimal number of unsupervised topics that the Latent Dirichlet Allocation may define, a range of values were tested using the model generator, and the relevant coherence score calculated. Ideally, the number of topics should be minimised to prevent too many similar topics being created, therefore maximising coherence, but not so little, that distinct topics are inadvertently placed under an `umbrella' category. Figure \ref{cohscore} shows the trend in coherence score for an increasing number of topics. There is generally an increasing trend, understandably due to \textit{The Simpsons}' longevity, and the broad diversity in plotlines, particularly due to the real-world influence on the writing

In terms of pre-processing, the only difference here was the use of a lemmatiser and stemmer, both of which reduce a word to its simplest base form, \textit{e.g.} the words ``running" and ``ran" would both be reduced to ``run". Unfortunately, the stemmer does possess the disadvantage of shortening proper nouns, including character names, limiting interpretation of words. Possible spelling errors in the script, whether intentionally done as part of the dialogue, or not, may also limit the stemmer's ability.

\begin{figure}[H]
    \centering
    \includegraphics[width=\textwidth]{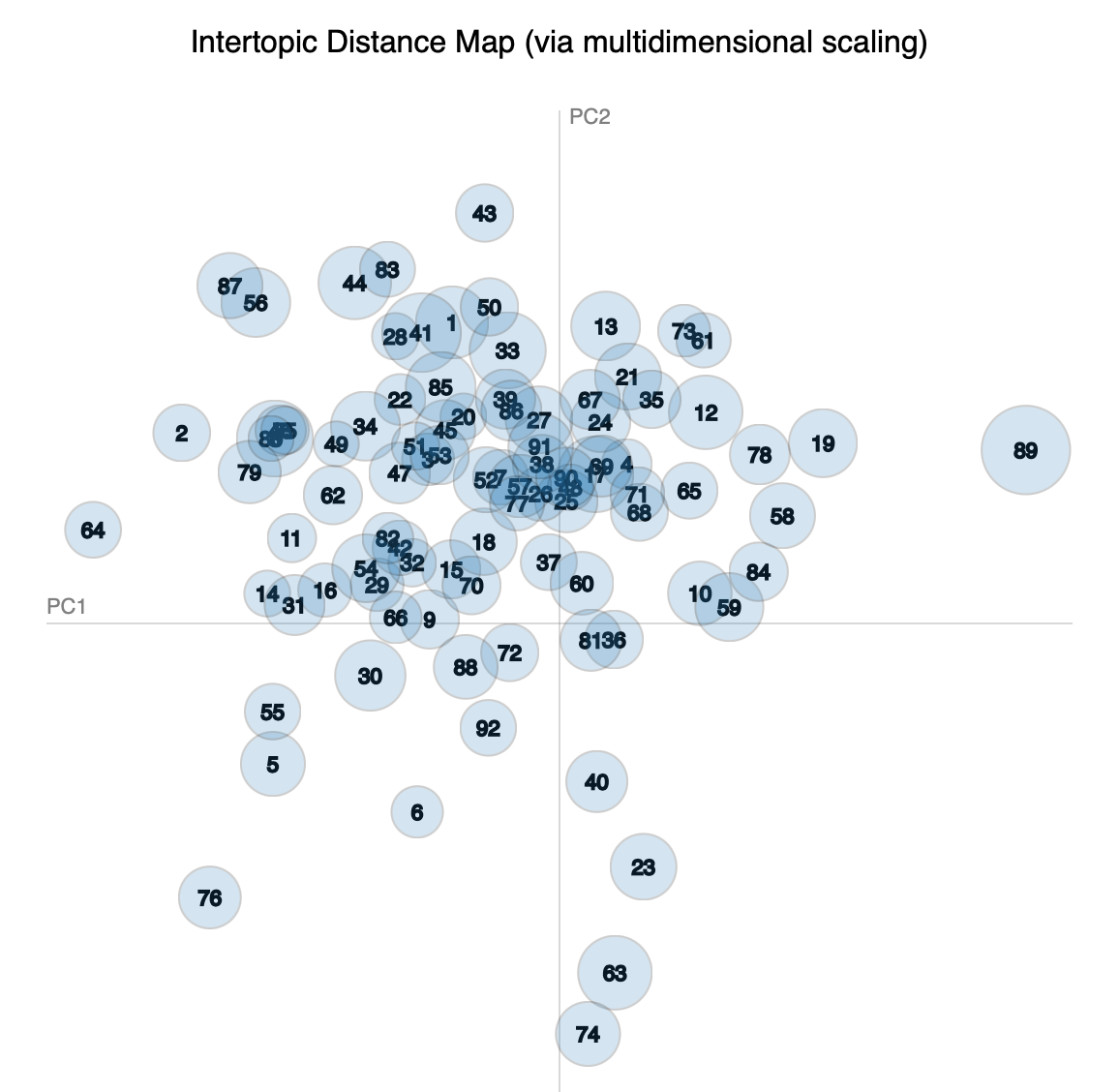}
    \caption{The distribution of the optimal $75$ topics.}
    \label{topicdist}
\end{figure}

Figure \ref{topicdist} shows the distribution of the topic bubbles, and their relative distances for the ideal number of topics, $75$. There is some overlap present around the centre, indicating topic similarity. However, the stemming does limit the interpretation of the topics. \texttt{top\_75\_topics.html} is linked on the Github repository, \href{https://github.com/athenax-coder/NLP-The-Simpsons-and-9-11}{NLP-The-Simpsons-and-9-11} for exploration.

\subsubsection{Supervised Allocation Using TF-IDF}
\begin{figure}[H]
    \centering
    \includegraphics[width=\textwidth]{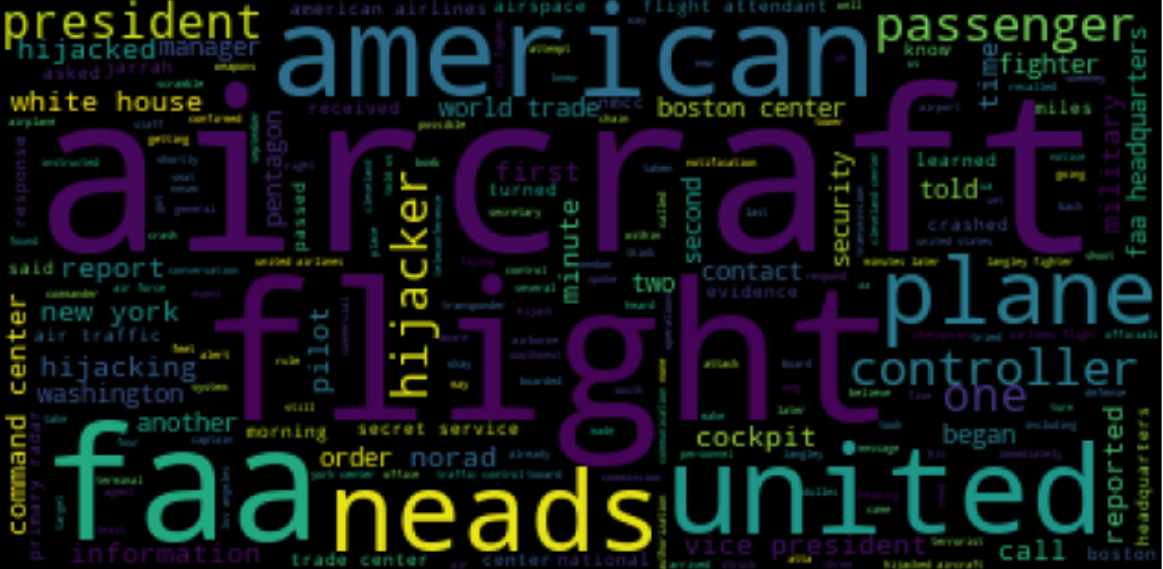}
    \caption{The word cloud for the $9/11$ commission report.}
    \label{cpwc}
\end{figure}

Since $9/11$ is a known topic of interest, performing supervised allocation of text is a logical step to take, however this requires having some known dictionary of words related to the topic. To find such words, similar tokenization and cleaning was performed on two sources of $9/11$ related text. Firstly, an address delivered by the then-president of the United States of America, George W. Bush, provided by \cite{wp_2001}. The other text used for comparison was the official government commission report published a few years after the attacks, entailing the major events before and after September $11$, $2001$, provided by \cite{kean_hamilton_2004}. Figure \ref{cpwc} shows the word cloud generated for the commission report, noting the common occurrences of words such as `terrorist' and `american'. 

The word ``applause" appears frequently in Bush's speech, seeing as it is an official transcript from the day of the address, which included the pauses allowed by Bush for the audience to react. In general, the Bush speech embodied more of the emotional connotations and patriotism associated with the general American public, whilst the commission report included many technical words related to the event, including flight details.

TF-IDF was used to properly weigh the importance of frequent words in both texts. \textit{Moby Dick} was used as the comparison, in keeping with the base text used in earlier comparisons. Notably, the age of the book may simply weigh more modern words, and upon replicating the procedure, perhaps a recent novel or even generic American government report could be used. For instance, the word `plane' may seldom appear in \textit{Moby Dick}, since it is a story largely revolving around boats. This would emphasise the frequency of `plane' in the 9/11 texts, even though a reference to a plane does not inherently refer to terrorism.

Upon doing so, the following correlation matrix was generated for the Bush speech and \textit{Moby Dick}, indicating an approximately $15.7\%$ correlation rate,
\[
\begin{bmatrix}
1 & 0.15675\\
0.15675 & 1
\end{bmatrix}.
\]
A similar rate of $13.3\%$ was seen between the commission report and \textit{Moby Dick}, most likely due to the higher content of political jargon,
\[
\begin{bmatrix}
1 & 0.13346\\
0.13346 & 1
\end{bmatrix}.
\]
Many of the words obtained in the TF-IDF frequency list were the same as those which appeared in the word cloud, again most likely due to very few of the common words in both texts being found in \textit{Moby Dick}. The top twenty-five and top fifty words from each were stored as lists, ready for use in comparison against \textit{The Simpsons}.

For every line in the cleaned dialogue, a counter variable was created, and added to if a unigram, that is, a single word or token in that sentence also appeared on the finalised list of 9/11-related words. This was done for the combined list of the top twenty-five words from the speech and report, as well as the combination of the top fifty from each.

\begin{figure}[H]
    \centering
    \includegraphics[width=\textwidth]{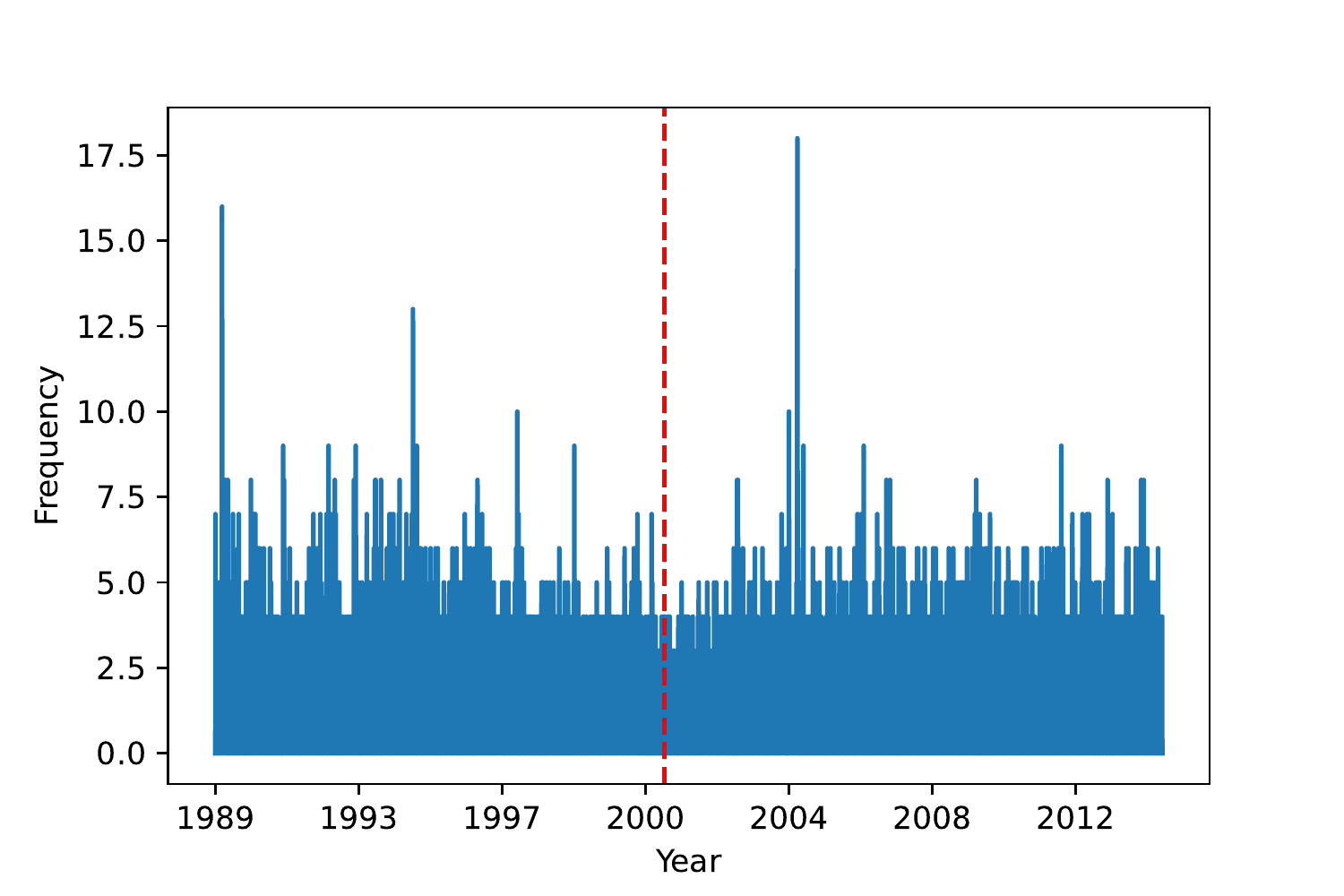}
    \caption{The frequencies of one-hundred total 9/11-related words in \textit{The Simpsons}. $2001$ is indicated by the red, dashed line.}
    \label{top50auto}
\end{figure}

\begin{figure}[H]
    \centering
    \includegraphics[width=\textwidth]{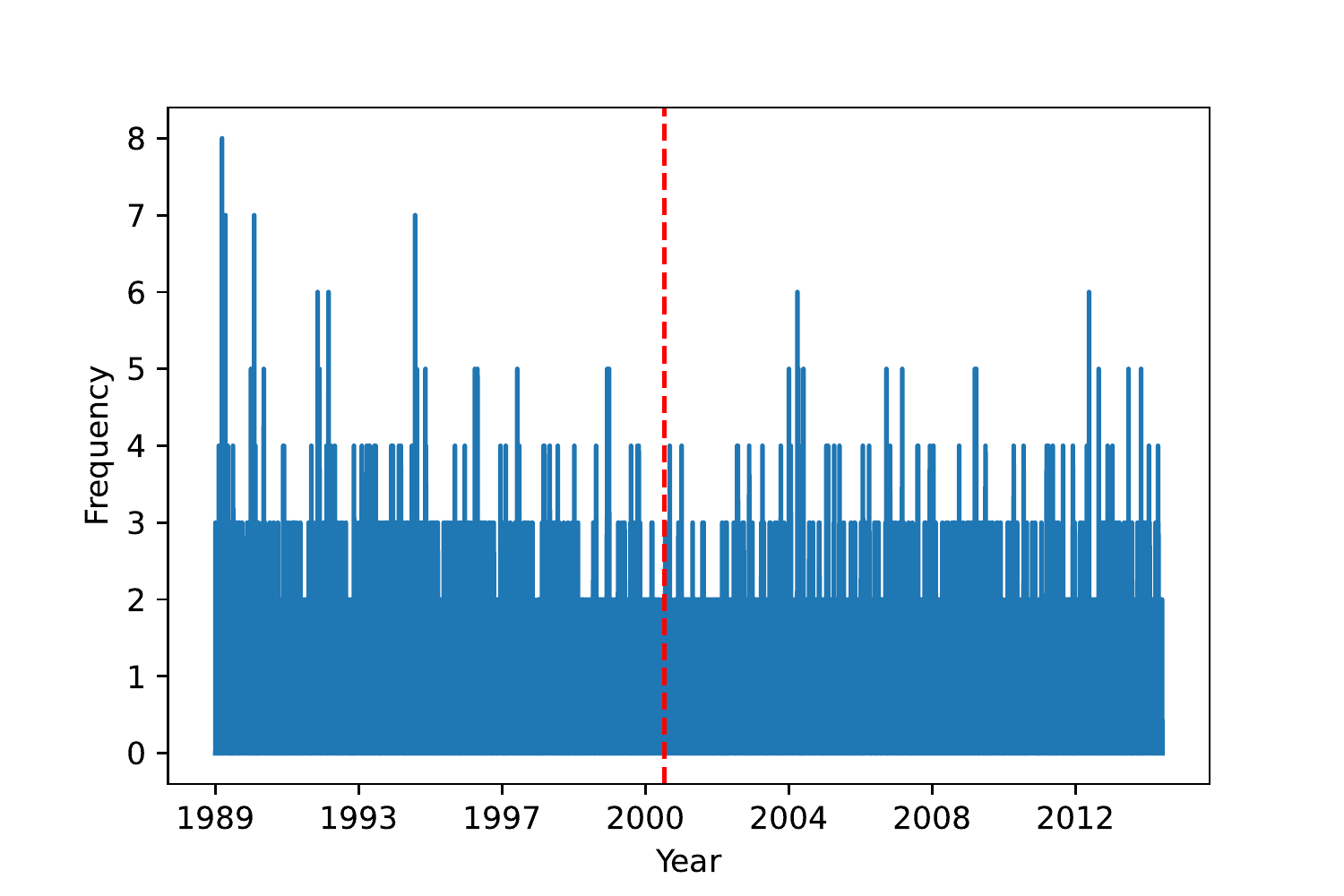}
    \caption{The frequencies of fifty total 9/11-related words in \textit{The Simpsons}. $2001$ is indicated by the red, dashed line.}
    \label{top25auto}
\end{figure}

Figure \ref{top50auto} and \ref{top25auto} show the frequency of the unigram commonality over time for the combined top fifty and top twenty-five list, respectively.

On average, there are around $0.335$ words per line for the larger list, and $0.150$ words per line for the shorter list. The fact that this frequency is nearly halved is logical considering the lengths of the lists are also halved.

However, there are words within these lists that may be removed on the basis of ambiguity or confidence in topic versatility. For example, the word `us', although likely referring to the abbreviation for the United States, `US', could also be counted when the pronoun `us' is used. Other generic words removed include,
\begin{multicols}{2}
\begin{itemize}
  \item `applause'
  \item `tonight'
  \item `world'
  \item `every'
  \item `us' 
  \item `many'
  \item `people'
  \item `country'
  \item `great'
  \item `ask'
  \item `come'
  \item `together'
  \item `seen'
  \item `thank'
  \item `may'
  \item `one'
  \item `strengthen'
  \item `asking'
  \item `want'
  \item `new'
  \item `forget'
  \item `lives'
  \item `end'
  \item `known'
  \item `center'
  \item `flight'
  \item `would'
  \item `plane'
  \item `minutes'
  \item `first'
  \item `told'
  \item `time'
  \item `call'
  \item `white'
  \item `house'
\end{itemize}
\end{multicols}

In total, $38$ words were removed from the top $100$ list, $15$ were removed from the top $50$ list. Some of these words also only possess meaning as bigrams, for example, `White House' refers to a government building in Washington, DC which does have relevance to 9/11, but `white' and `house' do not have meaning otherwise. Similarly, the trigram `World Trade Center' is a place very relevant to 9/11, but individually these words lack meaning.

\begin{figure}[H]
    \centering
    \includegraphics[width=\textwidth]{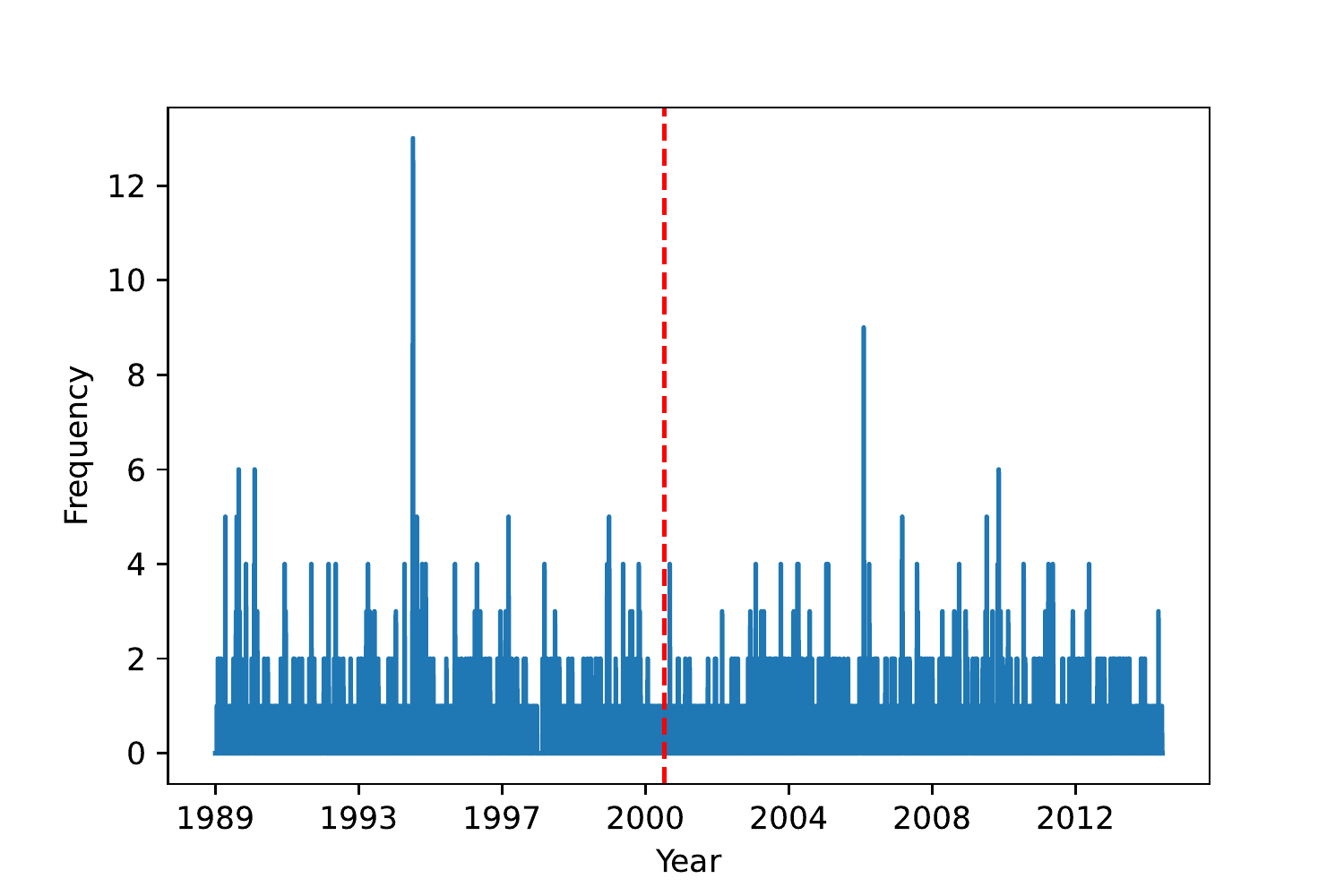}
    \caption{The frequencies of sixty-two specific 9/11-related words in \textit{The Simpsons}. $2001$ is indicated by the red, dashed line.}
    \label{top50manual}
\end{figure}

\begin{figure}[H]
    \centering
    \includegraphics[width=\textwidth]{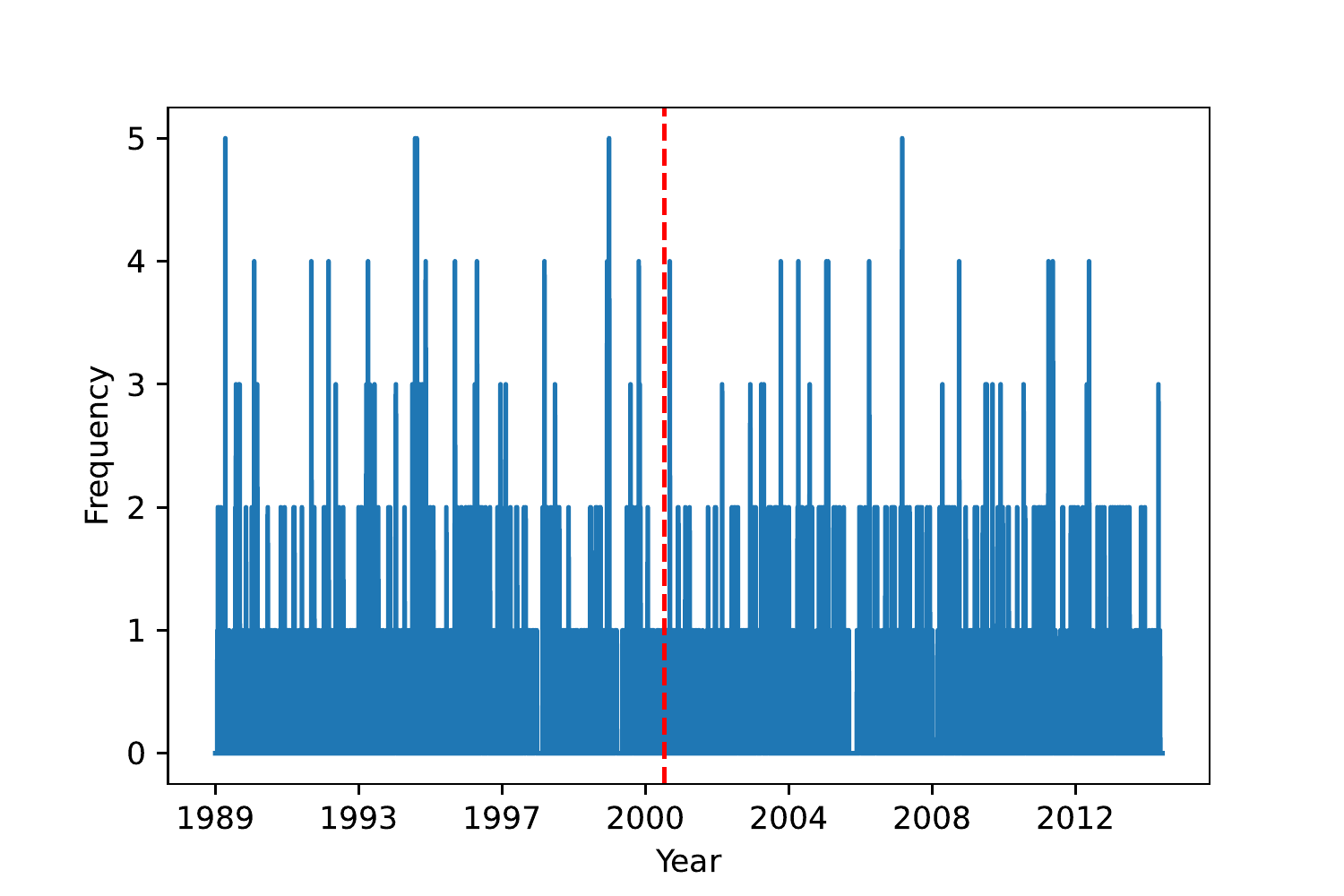}
    \caption{The frequencies of thirty-five specific 9/11-related words in \textit{The Simpsons}. $2001$ is indicated by the red, dashed line.}
    \label{top25manual}
\end{figure}

Figure \ref{top50manual} and \ref{top25manual} show the frequency of the unigram commonality over time for the combined top fifty and top twenty-five list, respectively, after manual filtering.

\vspace{1cm}
The average per line is now significantly reduced to $0.02335$ and $0.1514$ respectively. Generally, there is less noise in the data from the larger list in Figure \ref{top50manual}, with distinct peaks at around $1994$ and $2006$ being clear. Upon sorting the data frame by frequency, these peaks were found to be the lines ``fight fight fight fight fight fight fight fight fight fight fight fight fight" and ``chanting fight fight fight fight fight fight fight fight fight etc", respectively. `Fight' is a word generally related to violence, and was therefore kept in the manual filtering, but it still may be too generic. When considering Figure \ref{top25manual}, there is no real semblance of distinct change in frequency over the duration of the series.

\clearpage
\section{Conclusion}
While no significant change before and after 9/11 is obvious for \textit{The Simpsons}, there is certainly more to investigate. Firstly, the corpus of documents used to build a 9/11 vocabulary can be expanded upon and improved. Searching for relevant bigrams and trigrams can also be helpful. The sentiment did show a general decrease over time, but not significantly. There are also limitations in the stemming process that may help to refine some of the data cleaning as well. An extension for the project would be to perform a similar analysis for investigating changes in sentiment, word frequency, \textit{etc.} for other social issues such as gender equality, LGBTQ+ righs, and racial equality. Other possible investigations would be to focus only on the dialogue of particular characters, specifically the main family members. Generally, the results here and all possible future findings, are achieved based on the unique impact of real-world events on the writing and production of the show. 

\clearpage
\bibliographystyle{agsm-pp.bst}
\bibliography{TSA911NLPBIB.bib}

\end{document}